%% file: main.tex
\documentclass[10pt,twocolumn,letterpaper]{article}

\usepackage{cvpr}              


\definecolor{cvprblue}{rgb}{0.21,0.49,0.74}
\usepackage[pagebackref,breaklinks,colorlinks,allcolors=cvprblue]{hyperref}
\usepackage{amsmath,amsfonts}
\usepackage{algorithmic}
\usepackage{algorithm}
\usepackage{array}
\usepackage{textcomp}
\usepackage{stfloats}
\usepackage{url}
\usepackage{verbatim}
\usepackage{graphicx}
\usepackage{amsmath}
\usepackage{amsfonts}
\usepackage{textcomp}
\usepackage{xcolor}
\usepackage{booktabs}
\usepackage{multirow}
\usepackage{adjustbox}
\usepackage{color}
\usepackage{enumitem}
\usepackage{bbm}
\usepackage{bm} 
\usepackage{amsthm}
\usepackage{enumerate}
\usepackage{utfsym}
\usepackage{tikz-cd} 
\usepackage[accsupp]{axessibility}
\newcommand{\MyMapTemplatePrefix}[4]{\expandafter#1\csname#3#4\endcsname{#2{#4}}}
\newcommand{\MyMapTemplatePrefixNew}[5]{\expandafter#1\csname#4#5\endcsname{#2{#3{#5}}}}
\forcsvlist{\MyMapTemplatePrefix {\def} {\mathbf} {}} {A,B,C,D,E,F,G,H,I,J,K,L,M,N,O,P,Q,R,S,T,U,V,W,X,Y,Z}
\forcsvlist{\MyMapTemplatePrefix {\def} {\mathbf} {}} {a,b,c,d,e,f,g,h,i,j,k,l,m,n,o,p,q,r,s,t,u,v,w,x,y,z,1,0}
\forcsvlist{\MyMapTemplatePrefix {\def} {\widetilde} {wt}} {A,B,C,D,E,F,G,H,I,J,K,L,M,N,O,P,Q,R,S,T,U,V,W,X,Y,Z}
\forcsvlist{\MyMapTemplatePrefix {\def} {\widetilde} {wt}} {a,b,c,d,e,f,g,h,i,j,k,l,m,n,o,p,q,r,s,t,u,v,w,x,y,z} 
\forcsvlist{\MyMapTemplatePrefixNew {\def} {\widetilde}{\mathbf} {tb}} {A,B,C,D,E,F,G,H,I,J,K,L,M,N,O,P,Q,R,S,T,U,V,W,X,Y,Z}
\forcsvlist{\MyMapTemplatePrefixNew {\def} {\widetilde}{\mathbf} {tb}} {a,b,c,d,e,f,g,h,i,j,k,l,m,n,o,p,q,r,s,t,u,v,w,x,y,z}
\forcsvlist{\MyMapTemplatePrefix {\def} {\widehat} {wh}} {A,B,C,D,E,F,G,H,I,J,K,L,M,N,O,P,Q,R,S,T,U,V,W,X,Y,Z}
\forcsvlist{\MyMapTemplatePrefix {\def} {\widehat} {wh}} {a,b,c,d,e,f,g,h,i,j,k,l,m,n,o,p,q,r,s,t,u,v,w,x,y,z}
\forcsvlist{\MyMapTemplatePrefixNew {\def} {\widehat}{\mathbf} {hb}} {A,B,C,D,E,F,G,H,I,J,K,L,M,N,O,P,Q,R,S,T,U,V,W,X,Y,Z}
\forcsvlist{\MyMapTemplatePrefixNew {\def} {\widehat}{\mathbf} {hb}} {a,b,c,d,e,f,g,h,i,j,k,l,m,n,o,p,q,r,s,t,u,v,w,x,y,z}
\forcsvlist{\MyMapTemplatePrefix {\def} {\mathcal}{mc}} {A,B,C,D,E,F,G,H,I,J,K,L,M,N,O,P,Q,R,S,T,U,V,W,X,Y,Z}
\forcsvlist{\MyMapTemplatePrefix {\def} {\mathbb} {mb}} {A,B,C,D,E,F,G,H,I,J,K,L,M,N,O,P,Q,R,S,T,U,V,W,X,Y,Z}
\forcsvlist{\MyMapTemplatePrefix {\DeclareMathOperator} {} {} } {tr,diag,sgn}
\def\tp{^\intercal} \def\st{\text{s.t.~}}
\def\ie{\emph{i.e.}} 
 \def\eg{\emph{e.g.}}
\def\tr{\text{tr}}  \def\te{\text{te}}  \def\norm{\texttt{n}}  \def\anom{\texttt{a}} \def\res{\texttt{r}}
\def\bmu{\bm{\mu}} \def\bsigma{\bm{\sigma}}
\def\paperID{*****} 
\def\confName{CVPR}
\def\confYear{2025}

\title{Distribution Prototype Diffusion Learning for Open-set Supervised  \\ Anomaly Detection}

\author{
	Fuyun Wang\textsuperscript{1}, Tong Zhang\textsuperscript{1}\thanks{Corresponding authors: Tong Zhang,~Zhen Cui}, Yuanzhi Wang\textsuperscript{1}, 
	Yide Qiu\textsuperscript{1}, Xin Liu\textsuperscript{2}, Xu Guo\textsuperscript{1}, Zhen Cui\textsuperscript{3}\footnotemark[1] \\
	\textsuperscript{1}Nanjing University of Science and Technology, China. \\
	\textsuperscript{2}Nanjing SeetaCloud Technology, China.~
	\textsuperscript{3}Beijing Normal University, China
}

\begin{document}
\maketitle
\begin{abstract}
	In Open-set Supervised Anomaly Detection (OSAD), the existing methods typically generate pseudo anomalies to compensate for the scarcity of observed anomaly samples, while overlooking critical priors of normal samples, leading to less effective discriminative boundaries. To address this issue, we propose a Distribution Prototype Diffusion Learning (DPDL) method aimed at enclosing normal samples within a compact and discriminative distribution space. Specifically, we construct multiple learnable Gaussian prototypes to create a latent representation space for abundant and diverse normal samples and learn a Schrödinger bridge to facilitate a diffusive transition toward these prototypes for normal samples while steering anomaly samples away. Moreover, to enhance inter-sample separation, we design a dispersion feature learning way in hyperspherical space, which benefits the identification of out-of-distribution anomalies. Experimental results demonstrate the effectiveness and superiority of our proposed DPDL, achieving state-of-the-art performance on 9 public datasets.
\end{abstract}

\vspace{-20pt}
\section{Introduction}
Anomaly detection (AD)~\cite{kim2023sanflow, zhou2023anomalyclip,yao2024hierarchical,li2024promptad} aims to identify outliers significantly diverging from the prevailing samples in a dataset, and has a wide range of applications like industrial inspection, medical image analysis, and scientific discovery, etc. Recently, unsupervised anomaly detection (UAD)~\cite{dai2024generating, liu2024unsupervised,liu2024dual,he2024learning} and few-shot anomaly detection (FSAD)~\cite{liao2024coft, li2024promptad, hu2024anomalydiffusion} have emerged as prominent research paradigms, emphasizing the modeling of normal sample distributions to discern anomalies effectively. Yet, these methods often neglect prior knowledge from limited anomaly samples, resulting in imprecise delineation of normal sample boundaries and reduced efficacy in differentiating normal from anomaly instances. On the contrary, supervised anomaly detection (SAD)~\cite{li2023efficient, yao2023explicit, baitieva2024supervised} leverages a limited subset of anomaly samples as prior knowledge, improving detection performance. However, this reliance on  seen anomalies poses a risk of overfitting and hampers generalization to unseen anomalies in real-world settings.

To mitigate the challenge of limited generalization inherent in closed-set training, we focus on open-set supervised anomaly detection (OSAD)~\cite{liu2019margin, acsintoae2022ubnormal, zhu2022towards, zhu2024anomaly}, which utilizes a small set of known anomaly classes during training to identify unseen anomalies from open-set classes. By leveraging prior knowledge from observed samples, OSAD methods could reduce false positive errors. To improve the generalized detection of unseen anomalies, DRA~\cite{ding2022catching} leverages data augmentation and outlier exposure to learn a decomposed anomaly representation comprising seen anomalies, pseudo-anomalies, and potential residual anomalies. BGAD~\cite{yao2023explicit} leverages decision boundaries derived from normalized flow models to  capture and model anomaly information. Recently, AHL~\cite{zhu2024anomaly}  simulates heterogeneous anomaly distributions and performs collaborative differentiable learning to  further enhance the model's generality.

While data augmentation and outlier exposure techniques~\cite{pang2019deep,li2021cutpaste,ding2022catching} have demonstrated considerable success in anomaly detection, they fall short in generating comprehensive pseudo anomalies, capturing only a fraction of potential unseen anomalies. This limitation arises from overlooking the intricate nature of real-world anomaly distributions, thereby hindering the model's ability to generalize to novel anomaly types. Although AHL~\cite{zhu2024anomaly} has made strides in addressing this issue by simulating heterogeneous anomaly distributions, it still relies on approximating unknown out-of-distribution anomalies using known in-distribution anomalies for generalization. Yet, three critical issues persist: i) the simulation mechanism cannot cover all anomaly distribution patterns due to the varied scales and structures of anomaly distributions; ii) simulated anomalies inherit in-distribution data biases, leading to suboptimal performance on out-of-distribution anomalies; iii) the diversity of normal samples presents dilemmas for existing methods, complicating the differentiation between normal and anomaly boundaries. These issues prompt a fundamental question: 
\textit{Instead of generating pseudo and uncertain anomaly samples, how can we accurately  characterize compact distribution boundaries amidst a range of diverse normal samples and achieve robust generalization for unknown out-of-distribution anomalies?}

To address the aforementioned issue, in this work, we propose a Distribution Prototype Diffusion Learning (DPDL) method for open-set supervised anomaly detection. Considering abundant and diverse normal samples but very limited anomaly data, our method involves learning latent distribution prototypes, specifically multiple Gaussian distributions, onto which all observed normal samples can be effectively projected. To facilitate the mapping of normal samples into the prototype space, we leverage Schrödinger bridge (SB) framework, which enables a diffusive transition by aligning the distributions of these samples with the prototypes. The SB-based diffusion way could mitigate the out-of-distribution issue for normal samples to some extent. Within the distribution prototype space, we push observed anomaly samples away from normal samples to enhance discriminative capacity. Notably, both the prototypes and the diffusive bridge are learned jointly, resulting in a robust embedding space for normal samples. 
Moreover, to enhance generalization across unseen anomaly domains, we introduce a dispersion feature learning mechanism that maps intermediate features to a hyperspherical space, leveraging a mixture of von Mises-Fisher (vMF) distributions. This approach bolsters directional feature extraction and promotes robust inter-sample separation, facilitating effective identification of out-of-distribution samples. Experiments demonstrate that our method greatly improves detection capabilities for unseen anomalies.

In summary, our contributions are three-fold: i) we propose a distribution prototype diffusion learning framework that jointly learns multiple Gaussian prototypes and the associated diffusion bridge, creating a compact and discriminative embedding space; ii) we develop dispersion feature learning in hyperspherical space to enhance inter-sample separation and improve generalization; iii) we achieve state-of-the-art performance in 9 public datasets, demonstrating the efficacy of our approach.

\section{Related Work}

\textbf{Open-set Supervised Anomaly Detection.} Open-set supervised anomaly detection (OSAD) seeks to develop a robust anomaly detection framework that generalizes from a limited set of training anomalies to effectively identify previously unseen anomalies within an open-set context~\cite{liu2019margin, pang2021explainable, acsintoae2022ubnormal, zhu2022towards, yao2023explicit}.  Leveraging the prior knowledge provided by observed anomalies, contemporary OSAD methods significantly mitigate false positive errors, thereby enhancing overall detection performance~\cite{ding2022catching,zhu2024anomaly}. Recently, DRA~\cite{ding2022catching} learns disentangled representations of observed, pseudo, and residual anomalies to boost the detection of both seen and unseen anomalies. In contrast, AHL~\cite{zhu2024anomaly} simulates diverse heterogeneous anomaly distributions and employs collaborative differentiable learning, significantly improving the model’s generalization capacity. 


\textbf{Schrödinger Bridge.} Schrödinger bridge (SB), widely recognized as the entropy-regularized optimal transport (OT) problem, involves learning a stochastic process that evolves from an initial probability distribution to a target distribution under the influence of a reference measure~\cite{leonard2013survey, chen2016relation, de2021diffusion, liu2022deep,shi2024diffusion,liu2023generalized,kim2024fast}, which is often used for content generation as those prior generative models do~\cite{dicmor,imder,tcve,mmm-rs,reatco}.
$\text{I}^2$-SB~\cite{liu20232} and UNSB~\cite{kim2023unpaired} learn a nonlinear diffusion process between two given distributions or represent the SB problem as a series of adversarial learning problems to realize the image transformation task. Recently, LightSB~\cite{korotin2023light,gushchin2024light} introduces a novel, fast, and simple SB solver, which achieves optimal matching in practice through the Gaussian mixture parameterization of the adjusted schrödinger potential. 



\section{Preliminaries}

We focus on how to build the connection between two distributions $p_0$ and $p_1$, where the distributions are defined as absolutely continuous Borel probability distributions with finite second-order moments. Building upon the foundation of entropy-regularized optimal transport (EOT)~\cite{leonard2013survey, chen2016relation, de2021diffusion,korotin2023light,gushchin2024light, liu2022deep,noble2024tree,shi2024diffusion,liu2023generalized,kim2024fast}, we review the related properties of EOT and the schrödinger bridge (SB) problem with a Wiener prior. 


\textbf{Entropy-regularized optimal transport (EOT).} Given two point sets $\mcZ_0$ and $\mcZ_1$,   we seek for the optimal transport cost between any two points $z_0\in\mcZ_0$ and $z_1\in\mcZ_1$. This task may be formulated as an EOT problem with a parameter $\epsilon > 0$, \ie, minimizing the following objective:
\begin{equation}\label{eqn:eot}
	\begin{split}
		\mathop{\text{min}}\limits_{\pi \in \Pi (p_0,p_1)}\{\mathop{\int}\limits_{\mathbb{R}^D}\mathop{\int}\limits_{\mathbb{R}^D} \frac{1}{2} \Vert z_0 - z_1\Vert^2 \pi(z_0,z_1)dz_0dz_1 \\ + \epsilon \text{KL}(z\parallel p_0 \times p_1)\},
	\end{split}
\end{equation}
where $\Pi(p_0,p_1)$ denotes the set of transport plans, \ie, joint probability distributions on $\mathbb{R}^D \times \mathbb{R}^D$ with marginals $p_0$ and $p_1$, respectively, and KL denotes Kullback-Leibler divergence. The minimizer $\pi^*$ of Eqn.~(\ref{eqn:eot}) is guaranteed to exist, be unique, and absolutely continuous, and is referred as the EOT plan.

\textbf{Schrödinger Bridge (SB).} 
We define $\Omega$ as the space of $\mathbb{R}^D$-valued functions over time \(t \in [0, 1]\), representing trajectories in $\mathbb{R}^D$ that start at \(t = 0\) and end at \(t = 1\). We denote the set of probability distributions over \(\Omega\), i.e., stochastic processes, by \(\mathcal{P}(\Omega)\). The differential of the standard Wiener process is represented by \(dW_t\). For a process \(T \in \mathcal{P}(\Omega)\), we denote its joint distribution at \(t = 0, 1\) by \(\pi_T \in \mathcal{P}(\mathbb{R}^D \times \mathbb{R}^D)\). Similarly, we use \(T|_{z_0,z_1}\) to denote the distribution of \(T\) for \(t \in (0, 1)\), conditioned on \(T\)’s values \(z_0\) and \(z_1\) at \(t = 0\) and \(t = 1\), respectively.

Let \( W^\epsilon \in P(\Omega) \) represents a Wiener process with volatility \( \epsilon > 0 \), starting from \( p_0 \) at \( t = 0 \). Its differential is governed by the stochastic differential equation (SDE): \( dW^\epsilon_t = \sqrt{\epsilon} \, dW_t \). The Schrödinger bridge problem with the Wiener prior \( W^\epsilon \) between \( p_0 \) and \( p_1 \) is minimizing:
\begin{equation}\label{eq2}
	\mathop{\text{min}}\limits_{T \in \mathcal{F} (p_0,p_1)} \text{KL}(T\parallel W^\epsilon),
\end{equation}
where \( \mathcal{F}(p_0, p_1) \subset \mathcal{P}(\Omega) \) denotes the subset of stochastic processes that begin with distribution \( p_0 \) at \( t = 0 \) and reach \( p_1 \) at \( t = 1 \). This problem has a unique solution, a SDE diffusion process \( T^* \) defined: \( dZ_t = g^*(Z_t, t) \, dt + dW^\epsilon_t \). The process \( T^* \) is referred to as SB, and \( g^* : \mathbb{R}^D \times [0, 1] \rightarrow \mathbb{R}^D \) is the optimal drift.

\textbf{Characterization of solutions.} 
The EOT plan $\pi^*=\pi^{T^*}$ takes a specific form~\cite{korotin2023light, gushchin2024light}:
\begin{equation}\label{eq3}
	\pi^*(z_0,z_1) = u^*(z_0)\text{exp}(\frac{-\Vert z_0 - z_1 \Vert^2}{2\epsilon})v^*(z_1),
\end{equation}
where $u^*$, $v^*$: $\mathbb{R} \rightarrow \mathbb{R}_+$ are measurable functions known as Schrödinger potentials. The optimal drift \( g^* \) is derived as:
\begin{equation}
	g^*(z,t) = \epsilon\triangledown_z\text{log}\int_{\mathbb{R}^D}\mathcal{N}(z'|z,(1-t)\epsilon I_D)v^*(z')dz',\label{eqn:g}
\end{equation}



\section{Method}

\subsection{Problem Formulation}

Let $\mcX_\tr=\{(\texttt{x}_i,y_i)\}$ denotes a weakly-supervised training set with only image-level labels, where $\texttt{x}_i$ denotes one RGB image and $y_i\in\{0,1\}$ denotes whether $\texttt{x}_i$ is an anomaly sample (anomaly: $y_i = 1$, normal: $y_i = 0$). Hereby, $\mcX_\tr$ is consist of a normal subset $\mcX_\tr^\norm$ ($|\mcX_\tr^\norm|=N$) and an anomaly subset $\mcX_\tr^\anom$ ($|\mcX_\tr^\anom|=M$), formally, $\mcX_\tr\doteq\mcX_\tr^\norm\cup\mcX_\tr^\anom$, where generally $N\gg M$. Given a testing set $\mcX_\te$, we need to predict whether one sample $\texttt{x}\in\mcX_\te$ is anomaly or normal.
In OSAD, the anomaly patterns of the testing set do less recurred in those encountered training set. In other word, the distribution of anomalies are obviously discrepant, \ie, $P(\mcX_\te^\anom)\neq P(\mcX_\tr^\anom)$. Hereby, we need to learn a robust anomaly detection model $\psi$ from the training set $\mcX_\tr$, so that $\psi$ accurately infers anomaly scores for test samples. 

\begin{figure*}[htbp]
	\centering
	\includegraphics[scale=0.62]{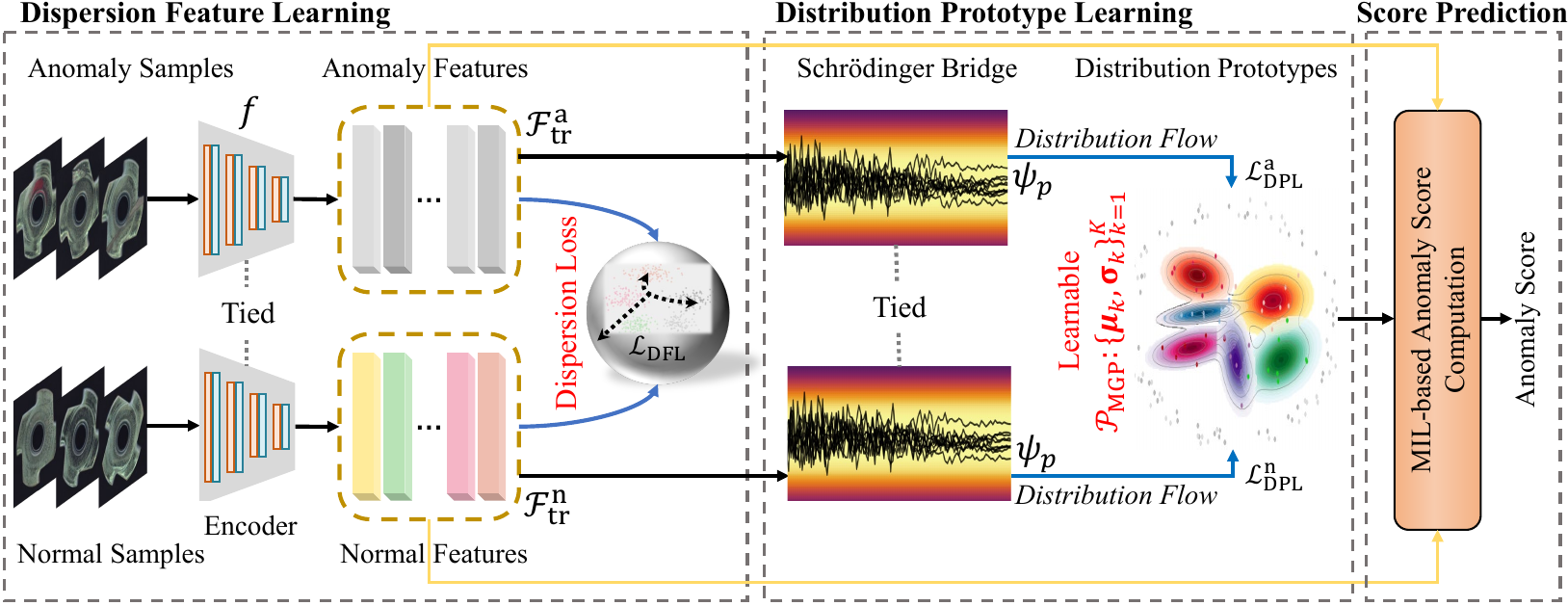}
	\caption{Our proposed DPDL framework. It comprises three distinct modules: Distribution Prototype Learning  (DPL, Sec.~\ref{sec:DPL}), Dispersion Feature Learning (DFL, Sec.~\ref{sec:DFL}), and anomaly score prediction (Sec.~\ref{sec:ASP}). DPL transforms the distribution of normal samples to a space of learnable multiple Gaussian prototypes through building schrödinger bridge, meantime pushing anomaly distribution away from these prototypes. DFL operates in a hyperspherical space, enlarging the distances of intermediate features of all samples in a hyperspherical space to strengthen feature generalization for detecting anomalies. The score prediction module leverages a multi-instance-learning method to compute anomaly scores.}
	\vspace{-10pt}
	\label{fig:overview}
\end{figure*}

\textit{Our abstract idea} is to learn latent distribution prototypes that not only encapsulate normal samples in a concise manner but also discriminate against anomaly samples. Given the abundance of observed normal samples, one natural approach is to characterize the distribution $P(\mcX_\tr^\norm)$, where samples outside this distribution, $\texttt{x}\notin P(\mcX_\tr^\norm)$, would be awarded higher probabilities as anomalies. Considering the inherent diversity of normal samples, we endeavor to learn multiple simple distributions (\eg, Gaussians) as prototypes $\mcP_\text{MGP}$, named multi-Gaussian prototypes (MGP). To embed input data into the prototype space, we introduce a generative bridge model $\psi_p$ for distribution transformation. The more \textit{abstract} formulation is given as follows:
\begin{align}
	\!\!\!\underset{\psi_p, \mcP_\text{MGP},f}{\min} & \!\! D_p(\psi_p(\mcF_\tr^\norm), \psi_p(\mcF_\tr^\anom),\mcP_\text{MGP})\!+\!\lambda D_{s}(\mcF_\tr^\norm, \mcF_\tr^\anom),\label{Eqn:abs-loss}\\
	\st, &~\psi_p: P(\mcF)\overset{\text{bridge}}{\longrightarrow} \mcP_\text{MGP}, \label{Eqn:abs-bridge}\\
	&~ f: \mcX\overset{\text{feature}}{\longrightarrow}\mcF,\label{Eqn:abs-disp}
\end{align}
where $\mcP_\text{MGP}$ denotes the distribution prototypes to be learned, $\psi_p$ is the flow function across probability distributions, $D_p$ signifies the discriminative function in the space of prototypes, $f$ stands for the feature extraction process, and $D_s$ acts as a regularizer to increase feature discriminability. In the above formulation, besides distribution prototype learning (DPL), we also introduce dispersion feature learning (DFL) executed within a hyperspherical embedding space, \ie, $D_s(\mcF_\tr^\norm, \mcF_\tr^\anom)$. 
The advantage of DFL is to prevent abrupt feature collapse in feature learning, thus preserving discriminative qualities. \textit{The overview of our framework} is abstractly depicted in Fig.~\ref{fig:overview}. Detailed elaboration on these aspects will be provided in the subsequent sections.

\subsection{Distribution Prototype Learning}\label{sec:DPL}

Distribution prototype learning is operated in the feature space. Hence, to extract intermediate image features, we can leverage those classic networks as the backbone, such as ResNet-18. Formally, the feature extraction process, denoted as a function $f:\mcX \rightarrow \mcF$,  transforms an image $\texttt{x}$ into the intermediate feature $\x=f(\texttt{x})\in\mbR^d$, where a flatten operation is used on convolution maps for vectorization. Consequently, we designate the intermediate feature sets of normal and anomaly samples as  $\mcF_\tr^\norm=\{\x_i^\norm|i=1,\cdots, N\}$ and $\mcF_\tr^\anom=\{\x_i^\anom|i=1,\cdots, M\}$, respectively. 

Considering the rarity of anomalies and the diversity of normal samples, it is pertinent to capture  the intricate distributions of normal samples by mapping the distribution $p_0=P(\mcF_\tr^\norm)$ of intermediate features to a distinct and well-characterized distribution $p_1$ like Gaussian. But due to various normal samples, we opt for a multi-Gaussian prototypes (MGP) comprising multiple Gaussian distributions as prototypes $\mcP_\text{MGP}=\{\mcP_i\doteq\mcN(\bmu_i,\bsigma_i)|i=1,\cdots, C\}$.
Given a normal sample $\x_i^\norm$, we expect to align it with the closest prototype with high likelihood. 
However, in the open-set setting, it is challenge to transform unseen points $\x \sim p_0$ to the target distribution $p_1$. Drawing inspiration from the capability of diffusion generation models in aligning disparate distributions, we frame the transition from source domain distribution $p_0$ to target domain distribution $p_1$ as a Schrödinger bridge problem. As formulated in Eqn.~(\ref{Eqn:abs-bridge}), we need learn an essential bridge flow $\psi_p: P(\mcF_\tr^\norm)\overset{\text{bridge}}{\longrightarrow} \mcP_\text{MGP}$. After bridge transformation, the condition probability in optimal transport plan conforms to: 
\begin{align}
	\pi(\psi_p(\x)|\x)\varpropto \underbrace{\sum^{C}_{c=1}\alpha_c\mathcal{N}(\psi_p(\x);\bmu_c, \bsigma_c))}_{\doteq\phi_1(\psi_p(\x))}, \label{eqn:GMM}
\end{align}
where the parameters $\{\alpha_c,\bmu_c,\bsigma_c\}_{c=1}^C$ (known as Gaussian mixed model (GMM), abstracted into the function $\phi_1$) as well as the flow function $\psi_p$ (in bridge) need to be learned. \textit{For simplicity, below we adopt diagonal matrices for $\bsigma_c$}.   

Drawing inspiration from previous works~\cite{korotin2023light, gushchin2024light}, we reframe distribution prototype learning in Eqn.~(\ref{eqn:GMM}) as the Schrödinger bridge. Given the specific form taken by the EOT plan described in Eqn.~(\ref{eq3}), we redefine these measurable functions $u, v: \mathbb{R} \rightarrow \mathbb{R}_+$, termed Schrödinger potentials~\cite{korotin2023light, gushchin2024light}, as follows:
\begin{align}
	u(\x^\norm_i) &\doteq \exp(\frac{\Vert \x_i^\norm\Vert^2}{2\epsilon})\phi_0(\x_i^\norm), \label{eqn:eq9}\\
	v(\psi(\x_i^\norm)) &\doteq \exp(\frac{\Vert \psi(\x_i^\norm)\Vert^2}{2\epsilon})\phi_1(\psi(\x_i^\norm)),  \label{eqn:eq10}
\end{align}
where $\phi_0$ is defined within the source feature domain $\mcF_\tr$, and $\phi_1$ is defined in Eqn.~(\ref{eqn:GMM}). $\epsilon$ is set to $0.001$ in our experiments. Accordingly, Eqn.~(\ref{eq3}) can be converted to:
\begin{align}\label{eqn:eq3}
	\!\!\!\! \pi(\x^\norm_i,\psi(\x_i^\norm)) \!=\!\! \phi_0(\x_i^\norm)\exp(\frac{\langle\x_i^\norm,\psi(\x_i^\norm)\rangle}{\epsilon})\phi_1(\psi(\x_i^\norm)),
\end{align}
Hence, the condition probability of the transport plan in Eqn.~(\ref{eqn:GMM}) could be exactly defined as
\begin{align}
	\pi(\psi(\x_i^\norm)|\x^\norm_i)\doteq\eta(\x_i^\norm,\psi(\x_i^\norm))\phi_1(\psi(\x_i^\norm)),
	\label{eqn:exact-GMM}
\end{align}
where the connection factor is denoted as $\eta(\x_i^\norm,\psi(\x_i^\norm))=\frac{1}{\varpi(\x_i^\norm)}\exp(\frac{\langle\x_i^\norm,\psi(\x_i^\norm)\rangle}{\epsilon})$ with the normalization term $\varpi(\x_i^\norm)=\int \exp(\langle \x_i^\norm,\psi(\x_i^\norm)\rangle)\phi_1(\psi(\x_i^\norm))d\psi(\x_i^\norm)$.
According to Eqns.~(\ref{eqn:GMM}) and (\ref{eqn:exact-GMM}), we can further derive a more tractable form~\cite{korotin2023light, gushchin2024light}:
\begin{align}
	&\!\!\!\!\pi(\psi(\x_i^\norm)|\x_i^\norm) \!\!=\!\! \widetilde{\eta}(\x_i^\norm)\!\!\sum^{C}_{c=1}\!\!  \widetilde{\alpha}_c(\x^\norm_i)\mcN(\psi(\x_i^\norm);\!\widetilde{\bmu}_c(\x_i^\norm), \!\bsigma_c),\!\!\label{eqn:pi-exact}
\end{align}
where the normalization factor is denoted as $\widetilde{\eta}(\x_i^\norm) = 1/\sum^{C}_{c=1}\widetilde{\alpha}_c(\x^\norm_i)$, the coefficients of multi-Gaussian are defined as $\widetilde{\alpha}_c(\x^\norm_i)= \alpha_c\!\exp(\frac{1}{2}(\x^\norm_i)\tp\bsigma_c\x^\norm_i+\frac{1}{\epsilon}(\widetilde{\bmu}_c)\tp(\x_i^\norm))$, and the mean vectors are calculated as $\widetilde{\bmu}_c(\x_i^\norm)=\bmu_c + \frac{1}{\epsilon}\bsigma_c\x_i^\norm$.


We proceed with deriving the bridge function $\psi_p$ that represents a SDE process: $d\x_t = g(x_t,t)dt + \sqrt{\epsilon} dW_t$ where the shift function $g$ is solved. According to Eqn.~(\ref{eqn:g}), we can obtain the shift function of diffusion process as follows:
\begin{align}
	\!\!\!\! g(\x_i^\norm,t) = \epsilon\rho_\text{MGP}(\x_i^\norm)\nabla_{\x_i^\norm}\log(\mcN(\x_i^\norm|0,\epsilon(1-t))\I),\label{eqn:g-gmm}
\end{align}
where the coefficients $\rho_\text{MGP}$ defined on multiple distribution prototypes is calculated as: $\rho_\text{MGP}(\x_i^\norm)=\sum_{c=1}^C(\alpha_c\mcN(\widetilde{\bmu}_c(\x_i^\norm)|\0,\bsigma_c)\mcN(h_c(\x_i^\norm,t)|0,\bm{\Sigma}_c^t))$, with another Gaussian of $h_c(\x_i^\norm,t) = \frac{1}{\epsilon(1-t)}\x_i^\norm + \bsigma^{-1}_c \widetilde{\bmu}_c(\x_i^\norm)$ and 
$\bm{\Sigma}_c^t = \frac{t}{\epsilon(1-t)}\I + \bsigma_c^{-1}$.


\textit{Distribution prototypes initialization.} Jointly learning prototypes $\{\alpha_c, \bmu_c, \bsigma_c\}_{c=1}^C$ and the bridge transformation $\psi_p$ (also the shift $g$) is a challenging task as they are interdependent. For this, we leverage a vector quantization function to learn a codebook $\mcE$ of prototypes within a discrete latent space from training data. Specifically, given an input image feature $\x_i^\norm$, we can assign $\x_i^\norm$ to the closest prototype $\e_k$ by minimizing the L2 distance between $\x_i^\norm$ and each of prototypes $\e_c\in\mcE$, as follows:
\begin{align}
	\!\!\!\!\underset{\{\e_c\}}{\min}~\mbE_{\x_i^\norm\in\mcF_\tr}[\Vert \x_i^\norm \!\!-\!\e_{c^*} \Vert^2_2], \st, c^* \!\!= \!\!\underset{c}{\arg\min}\|\x_i^\norm\!\!-\!\e_c\|_2,\!\!
\end{align}
where $c^*$ denotes the index of the prototype closest to $\x_i^\norm$. The learned $\{\e_c\}_{c=1}^C$ are used to initialize the mean vectors $\{\bmu_c\}_{c=1}^C$, while the variances of all prototypes are set to the identity matrix. We observe that this initialization strategy accelerates the training process.


\textit{Distribution loss of normal and anomaly samples.} 
As \( p_0 \) and \( p_1 \) are accessible only via samples $\mcF_\tr$ and prototypes $\mcP_\text{MGP}$, we optimize the empirical form in Eqn.~(\ref{eq2}) for normal samples as follows~\cite{korotin2023light, gushchin2024light}:
\begin{equation}
	\mcL_\text{DPL}^\norm = \frac{1}{N}\sum^{N}_{i=1}\log \varpi_\theta(\x_i^\norm) - \frac{1}{C}\sum^{C}_{c=1}\log \phi_1(\bmu_c),
\end{equation}
In contrast, for anomaly samples, we aim to push anomaly distribution away from $p_1$, \ie, negative loss, formally,
\begin{equation}
	\mcL_\text{DPL}^\anom = \frac{1}{C}\sum^{C}_{c=1}\log \phi_1(\bmu_c)-\frac{1}{M}\sum^{M}_{i=1}\log \varpi_\theta(\x_i^\anom),
\end{equation}

\subsection{Dispersion Feature Learning}\label{sec:DFL}

A critical challenge in OSAD is detecting previously unseen anomalies in open-set environments. Due to the limited anomaly observations, existing methods often use pseudo-anomaly generation strategies as a means of effective data augmentation. Nonetheless, the effectiveness of these methods heavily depends on the quality of the pseudo-anomaly feature embeddings. Notably, pseudo-anomaly distributions often inherit biases from the in-distribution data, which differ from the unknown anomaly distributions in out-of-distribution data. Current methods fail to address the relationship between observed and unknown anomalies, particularly for out-of-distribution generalization.


To improve out-of-distribution detection, it is essential to promote a larger inter-sample dispersion, as greater distances among in-distribution samples facilitate their more effective separation from out-of-distribution samples. In other words, if all inter-sample distances are close to zero, \ie, collapse to a single point, it becomes impossible to differentiate between samples. Hence, promoting separability through a larger inter-sample dispersion is critical for accurately identifying samples that do not belong to known categories. To do so, we map the features into a \textit{hyperspherical} space. Draws inspiration from the vMF distribution~\cite{mardia2009directional} in directional statistics, we compute spherical Gaussian distributions for unit-norm features $\hbx_i = \x_i/\|\x_i\|_2^2$. The probability density function of a unit vector $\hbx_i \in \mathbb{R}^D$ is defined in the \textit{hyperspherical} space as follows:
\begin{equation}
	p_D(\hbx_i;\hbx_j,\kappa) = F_D(\kappa)\exp(\kappa\langle\hbx_i,\hbx_j\rangle),
\end{equation}
where \( \kappa \geq 0 \) controls the concentration of the distribution around the mean direction $\hbx_j$ and \( F_D(\kappa) \) is the normalization factor. A larger \( \kappa \) value increases concentration around the mean, while in the extreme case \( \kappa = 0 \), sample points are uniformly distributed on the hypersphere. Therefore, we design a dispersion loss to optimize large angular distances between the features of all samples:
\begin{equation}
	\!\! \!\!\mathcal{L}_\text{DFL} \!\!=\!\! \frac{1}{U}\!\!\sum^{U}_{i=1}\log\frac{1}{U-1}\!\!\sum_{i,j=1}^{U}\!\!\mathbbm{1}\{i \neq j\}\exp(\kappa\langle\hbx_i,\hbx_j\rangle), 
\end{equation}
where $U=N+M$, and $\kappa$ is set to $10$ in our experiments.


\subsection{Anomaly Score Prediction}\label{sec:ASP}

Based on the above designs, we leverage the multiple-instance-learning (MIL)-based method proposed in~\cite{pang2021explainable} to  effectively learn anomaly scores. Similar to the work~\cite{ding2022catching}, we design three modules $\mathsf{M}$ = \{$\mathsf{M}_\anom$, $\mathsf{M}_\norm$ and $\mathsf{M}_\res$\} for estimating anomaly scores. Firstly, for the feature map $\x_i$\footnote{Here $\x_i$ refers to convolution maps without flattening (as used above).}, we generate pixel-wise feature vectors $\mcV = \{\v_i\}_{i=1}^{H' \times W'} $ to represent the feature of small patches of the image $\texttt{x}_i$, where $(H',W')$ denotes the size of the feature map. 
These pixel-wise representations are then mapped by an anomaly classifier $S_\anom$ to estimate pixel-level anomaly scores. To capture those points with the most salient anomalies, we compute  the top-$K$ most anomaly pixel points and define the loss function as:
\begin{equation}\label{eq16}
	\mathcal{L}_{\mathsf{M}_\anom}(\x_i,y_i) = \mathcal{L}_\text{binary}
	(\frac{1}{K}{\sum}
	\text{Top}_K \{S_\anom(\v_i;\theta_\anom)\},y_i),
\end{equation}
where $\mathcal{L}_\text{binary}$ refers to a binary classification loss function, and $\text{Top}_K$ selects the highest $K$ anomaly scores among all the vectors. We use $\mathsf{M}_\norm$ to learn the normal features:
\begin{equation}
	\mathcal{L}_{\mathsf{M}_\norm}(\x_i,y_i) = \mathcal{L}_\text{binary}(S_\norm(\frac{1}{H'\times W'}\sum^{H'\times W'}_{i=1}\v_i;\theta_\norm),y_i), 
\end{equation}
where $S_\norm:\mcV \rightarrow \mathbb{R}$ is a fully connected binary anomaly classifier. Finally, we define $\mathsf{M}_\res$ to compute the residual 
anomaly scores between fine-grained visual semantics and abstract prototypes:
\begin{align}
	\mathcal{L}_{\mathsf{M}_\res} = \mathcal{L}_\text{binary}(S_\res((\psi_p(\x_i)-\bmu_{c^*})/\bsigma_{c^*};\theta_\res),y_i),
\end{align}
where $c^*$ denotes the index of the most probable prototypes, \ie, $c^*=\arg\max_c \mcN(\psi_p(\x_i);\bmu_c,\bsigma_c)$, and $S_\res$ utilize the same method to obtain anomaly score as $S_\anom$. 

\textbf{Training.}  During the training phase, the SB and three prediction modules are jointly trained. To this end, we employ an objective function that encompasses three components as follows:
\begin{align}
	\mathcal{L} = \underbrace{\mathcal{L}_{\mathsf{M}_\anom}+\mathcal{L}_{\mathsf{M}_\norm}+\mathcal{L}_{\mathsf{M}_\res}}_\text{MIL-based learning} + \underbrace{\mathcal{L}_\text{DPL}^\norm + \mathcal{L}_\text{DPL}^\anom}_\text{SB transform} + \underbrace{\lambda \mathcal{L}_\text{DFL}}_\text{dispersion}.
\end{align}
where the coefficient $\lambda$ modulates the relative importance  of dispersion loss, and the learnable parameters include $\{\alpha_c,\bmu_c,\bsigma_c\}$, $\theta_{\psi_p}$, and $\{\theta_\anom, \theta_\norm,\theta_\res\}$. 

\textbf{Inference.}
During the test phase, we find the most similar class prototype through SB, and subsequently compute the anomaly score by adding the scores from both \( S_\anom \) and \( S_\res \), while subtracting the normal score obtained from \( S_\norm\) for the given test image.


\section{Experiment}

\subsection{Dataset and Evaluation Metric}
\textbf{Dataset}
To validate the effectiveness of DPDL, comprehensive experiments are conducted on nine real-world AD datasets, including six industrial defect detection datasets (MVTec AD~\cite{bergmann2019mvtec}, Optical~\cite{wieler2007weakly}, SDD~\cite{tabernik2020segmentation}, AITEX~\cite{silvestre2019public}, ELPV~\cite{deitsch2019automatic}, Mastcam~\cite{kerner2020comparison}) and three medical image datasets (Hyper-Kvasir~\cite{borgli2020hyperkvasir}, Brain-MRI~\cite{salehi2021multiresolution}, HeadCT~\cite{salehi2021multiresolution}). We follow the previous OSAD baselines~\cite{ding2022catching,zhu2024anomaly} to adopt two protocols for sampling, including general setting and hard setting. The general setting assumes that anomaly examples are randomly sampled from the anomaly class, while the hard setting samples from a single class to assess generalization to new or unseen anomaly classes.

\noindent
\textbf{Evaluation Metric}
We utilize the widely adopted Area Under ROC Curve (AUC) as a metric to evaluate the performance across all methods and settings.  All reported AUCs are averaged results over five independent runs.

\begin{table*}[htbp]\small
	\belowrulesep=0pt
	\aboverulesep=0pt
	\centering
	\caption{ AUC performance (mean ± std) across nine real-world AD datasets is reported under the general setting. $\textcolor{red}{\textbf{red}}$ highlights the best results, and $\textcolor{blue}{\textbf{blue}}$ indicates sub-optimal outcomes. All baseline SOTA results are sourced from the original papers~\cite{ding2022catching,zhu2024anomaly}.}
	\renewcommand{\arraystretch}{1.0}
		\begin{tabular}{c|cccccc|c}
			\toprule
			\multirow{1}{*}{Dataset}  &  DevNet & FLOS & SAOE & MLEP & DRA & AHL & DPDL (Ours) \\
			\midrule
			\multicolumn{8}{c}{Ten  Training Anomaly Examples} \\
			\midrule
			\textbf{MVTec AD} &  0.945$\pm$0.004 & 0.939$\pm$0.007 & 0.926$\pm$0.010 & 0.907$\pm$0.005 & 0.959$\pm$0.003 & \textcolor{blue}{\textbf{0.970}}$\pm$0.002 & \textcolor{red}{\textbf{0.977}}$\pm$0.002  \\
			\textbf{Optical} &  0.782$\pm$0.065 & 0.720$\pm$0.055 & 0.941$\pm$0.013 & 0.740$\pm$0.039  & 0.965$\pm$0.006 & \textcolor{blue}{\textbf{0.976}}$\pm$0.004 & \textcolor{red}{\textbf{0.983}}$\pm$0.005   \\
			\textbf{SDD}  & 0.988$\pm$0.006 & 0.967$\pm$0.018 & 0.955$\pm$0.020 & 0.983$\pm$0.013   & 0.991$\pm$0.005 & \textcolor{blue}{\textbf{0.991}}$\pm$0.001 & \textcolor{red}{\textbf{0.996}}$\pm$0.001 \\
			\textbf{AITEX}   & 0.887$\pm$0.013 & 0.841$\pm$0.049 & 0.874$\pm$0.024 & 0.867$\pm$0.037  & 0.893$\pm$0.017 & \textcolor{blue}{\textbf{0.925}}$\pm$0.013 & \textcolor{red}{\textbf{0.975}}$\pm$0.007 \\
			\textbf{ELPV}   & 0.846$\pm$0.022 & 0.818$\pm$0.032 & 0.793$\pm$0.047 & 0.794$\pm$0.047  & 0.845$\pm$0.013 & \textcolor{blue}{\textbf{0.850}}$\pm$0.004 & \textcolor{red}{\textbf{0.937}}$\pm$0.003  \\
			\textbf{Mastcam}  & 0.790$\pm$0.021 & 0.703$\pm$0.029 & 0.810$\pm$0.029 & 0.798$\pm$0.026  & 0.848$\pm$0.008 & \textcolor{blue}{\textbf{0.855}}$\pm$0.005 & \textcolor{red}{\textbf{0.934}}$\pm$0.010  \\
			\textbf{Hyper-Kvasir}  & 0.829$\pm$0.018 & 0.773$\pm$0.029 & 0.666$\pm$0.050 & 0.600$\pm$0.069  & 0.834$\pm$0.004 & \textcolor{blue}{\textbf{0.880}}$\pm$0.003 & \textcolor{red}{\textbf{0.939}}$\pm$0.005 \\
			\textbf{BrainMRI}  & 0.958$\pm$0.012 & 0.955$\pm$0.011 & 0.900$\pm$0.041 & 0.959$\pm$0.011  & \textcolor{blue}{\textbf{0.970}}$\pm$0.003 & \textcolor{red}{\textbf{0.977}}$\pm$0.001 & 0.969$\pm$0.005  \\
			\textbf{HeadCT}  & \textcolor{blue}{\textbf{0.982}}$\pm$0.009 & 0.971$\pm$0.004 & 0.935$\pm$0.021 & 0.972$\pm$0.014  & 0.972$\pm$0.002 & \textcolor{red}{\textbf{0.999}}$\pm$0.003 & 0.981$\pm$0.003  \\
			\midrule
			
			\multicolumn{8}{c}{One Training Anomaly Example}  \\
			\midrule
			\textbf{MVTec AD} &  0.780$\pm$0.020 & 0.755$\pm$0.136 & 0.834$\pm$0.007 & 0.744$\pm$0.019 & 0.883$\pm$0.008 & \textcolor{blue}{\textbf{0.901}}$\pm$0.003 & \textcolor{red}{\textbf{0.927}}$\pm$0.002  \\
			\textbf{Optical} &  0.523$\pm$0.003 & 0.518$\pm$0.003 & 0.815$\pm$0.014 & 0.516$\pm$0.009  & 0.888$\pm$0.012 & \textcolor{blue}{\textbf{0.888}}$\pm$0.007 & \textcolor{red}{\textbf{0.915}}$\pm$0.002   \\
			\textbf{SDD}  & 0.881$\pm$0.009 & 0.840$\pm$0.043 & 0.781$\pm$0.009 & 0.811$\pm$0.045   & 0.859$\pm$0.014 & \textcolor{blue}{\textbf{0.909}}$\pm$0.001 & \textcolor{red}{\textbf{0.917}}$\pm$0.003 \\
			\textbf{AITEX}   & 0.598$\pm$0.070 & 0.538$\pm$0.073 & 0.675$\pm$0.094 & 0.564$\pm$0.055  & 0.692$\pm$0.124 & \textcolor{blue}{\textbf{0.734}}$\pm$0.008 & \textcolor{red}{\textbf{0.838}}$\pm$0.008 \\
			\textbf{ELPV}   & 0.514$\pm$0.076 & 0.457$\pm$0.056 & 0.635$\pm$0.092 & 0.578$\pm$0.062  & 0.675$\pm$0.024 & \textcolor{blue}{\textbf{0.828}}$\pm$0.005 & \textcolor{red}{\textbf{0.897}}$\pm$0.002  \\
			\textbf{Mastcam}  & 0.595$\pm$0.016 & 0.542$\pm$0.017 & 0.662$\pm$0.018 & 0.625$\pm$0.045  & 0.692$\pm$0.058 & \textcolor{blue}{\textbf{0.743}}$\pm$0.003 & \textcolor{red}{\textbf{0.838}}$\pm$0.011  \\
			\textbf{Hyper-Kvasir}  & 0.653$\pm$0.037 & 0.668$\pm$0.004 & 0.498$\pm$0.100 & 0.445$\pm$0.040  & 0.690$\pm$0.017 & \textcolor{blue}{\textbf{0.768}}$\pm$0.015 & \textcolor{red}{\textbf{0.821}}$\pm$0.007  \\
			\textbf{BrainMRI}  & 0.694$\pm$0.004 & 0.693$\pm$0.036 & 0.531$\pm$0.060 & 0.632$\pm$0.017  & 0.744$\pm$0.004 & \textcolor{blue}{\textbf{0.866}}$\pm$0.004 & \textcolor{red}{\textbf{0.893}}$\pm$0.004  \\
			\textbf{HeadCT}  & 0.742$\pm$0.076 & 0.698$\pm$0.092 & 0.597$\pm$0.022 & 0.758$\pm$0.038  & 0.796$\pm$0.105 & \textcolor{blue}{\textbf{0.825}}$\pm$0.014 & \textcolor{red}{\textbf{0.865}}$\pm$0.005  \\
			\bottomrule
		\end{tabular}
		\label{tab:main_performance}
	\end{table*}
	
	\begin{table*}[htbp]\small
		\belowrulesep=0pt
		\aboverulesep=0pt
		\centering
		\caption{AUC results (mean ± std) under the hard setting.  The best and second-best results are highlighted in $\textcolor{red}{\textbf{red}}$ and $\textcolor{blue}{\textbf{blue}}$, respectively.  Carpet and Metal\_nut are subsets of MVTec AD.  The datasets used are consistent with those in ~\cite{ding2022catching,zhu2024anomaly}, where those datasets only containing one anomaly class are excluded to adapt for the hard setting.}
		\setlength{\tabcolsep}{1.5mm}
		\renewcommand{\arraystretch}{1.0}
			\begin{tabular}{c|cccccc|c}
				\toprule
				\multirow{1}{*}{Dataset}  &  DevNet & FLOS & SAOE & MLEP & DRA & AHL & DPDL (Ours) \\
				\midrule
				\multicolumn{8}{c}{Ten  Training Anomaly Examples} \\
				\midrule
				\textbf{Carpet} \textcolor{gray}{\text{(mean)}} &  0.847$\pm$0.017 & 0.761$\pm$0.012 & 0.762$\pm$0.073 & 0.751$\pm$0.023 & 0.935$\pm$0.013 & \textcolor{blue}{\textbf{0.949}}$\pm$0.002 & \textcolor{red}{\textbf{0.956}}$\pm$0.004  \\
				\textbf{Metal\_nut} \textcolor{gray}{\text{(mean)}} &  0.965$\pm$0.011 & 0.922$\pm$0.014 & 0.855$\pm$0.016 & 0.878$\pm$0.058  & 0.945$\pm$0.017 & \textcolor{blue}{\textbf{0.972}}$\pm$0.002 & \textcolor{red}{\textbf{0.978}}$\pm$0.002   \\
				
				\textbf{AITEX}  \textcolor{gray}{\text{(mean)}}  & 0.683$\pm$0.032 & 0.635$\pm$0.043 & 0.724$\pm$0.032 & 0.626$\pm$0.041  & 0.733$\pm$0.009 & \textcolor{blue}{\textbf{0.747}}$\pm$0.002 & \textcolor{red}{\textbf{0.798}}$\pm$0.005 \\
				\textbf{ELPV} \textcolor{gray}{\text{(mean)}}   & 0.702$\pm$0.023 & 0.642$\pm$0.032 & 0.683$\pm$0.047 & 0.745$\pm$0.020  & 0.766$\pm$0.029 & \textcolor{blue}{\textbf{0.788}}$\pm$0.003 & \textcolor{red}{\textbf{0.818}}$\pm$0.003  \\
				\textbf{Mastcam} \textcolor{gray}{\text{(mean)}}   & 0.588$\pm$0.011 & 0.616$\pm$0.021 & 0.697$\pm$0.014 & 0.588$\pm$0.016  & 0.695$\pm$0.004 & \textcolor{blue}{\textbf{0.721}}$\pm$0.003 & \textcolor{red}{\textbf{0.778}}$\pm$0.007  \\
				\textbf{Hyper-Kvasir} \textcolor{gray}{\text{(mean)}}  & 0.822$\pm$0.019 & 0.786$\pm$0.021 & 0.698$\pm$0.021 & 0.571$\pm$0.014  & 0.844$\pm$0.009 & \textcolor{blue}{\textbf{0.854}}$\pm$0.004 & \textcolor{red}{\textbf{0.864}}$\pm$0.002 \\
				\midrule
				
				\multicolumn{8}{c}{One Training Anomaly Example}  \\
				\midrule
				\textbf{Carpet} \textcolor{gray}{\text{(mean)}}  &  0.767$\pm$0.018 & 0.678$\pm$0.040 & 0.753$\pm$0.055 & 0.679$\pm$0.029 & 0.901$\pm$0.006 & \textcolor{blue}{\textbf{0.932}}$\pm$0.003 & \textcolor{red}{\textbf{0.941}}$\pm$0.006  \\
				\textbf{Metal\_nut} \textcolor{gray}{\text{(mean)}}  &  0.855$\pm$0.016 & 0.855$\pm$0.024 & 0.816$\pm$0.029 & 0.825$\pm$0.023  & 0.932$\pm$0.017 & \textcolor{blue}{\textbf{0.939}}$\pm$0.004 & \textcolor{red}{\textbf{0.944}}$\pm$0.003   \\
				\textbf{AITEX} \textcolor{gray}{\text{(mean)}}   & 0.646$\pm$0.034 & 0.624$\pm$0.024 & 0.674$\pm$0.034 & 0.466$\pm$0.030  & 0.684$\pm$0.033 & \textcolor{blue}{\textbf{0.707}}$\pm$0.007 & \textcolor{red}{\textbf{0.753}}$\pm$0.005 \\
				\textbf{ELPV} \textcolor{gray}{\text{(mean)}}   & 0.648$\pm$0.057 & 0.691$\pm$0.008 & 0.614$\pm$0.048 & 0.566$\pm$0.111  & 0.703$\pm$0.022 & \textcolor{blue}{\textbf{0.740}}$\pm$0.003 & \textcolor{red}{\textbf{0.762}}$\pm$0.003  \\
				\textbf{Mastcam} \textcolor{gray}{\text{(mean)}}  & 0.511$\pm$0.013 & 0.524$\pm$0.013 & 0.689$\pm$0.037 & 0.541$\pm$0.007  & 0.667$\pm$0.012 & \textcolor{blue}{\textbf{0.673}}$\pm$0.010 & \textcolor{red}{\textbf{0.733}}$\pm$0.004  \\
				\textbf{Hyper-Kvasir} \textcolor{gray}{\text{(mean)}}   & 0.595$\pm$0.023 & 0.571$\pm$0.004 & 0.406$\pm$0.018 & 0.480$\pm$0.044  & 0.700$\pm$0.009 & \textcolor{blue}{\textbf{0.706}}$\pm$0.007 & \textcolor{red}{\textbf{0.715}}$\pm$0.004  \\
				\bottomrule
			\end{tabular}
			\label{tab:hard_setting}
						\vspace{-10pt}
		\end{table*}

		\subsection{Baselines}
		We compare DPDL against six related state-of-the-art OSAD baselines, including SAOE~\cite{markovitz2020graph,tack2020csi,li2021cutpaste}, MLEP~\cite{liu2019margin}, FLOS~\cite{ross2017focal}, DevNet~\cite{pang2021explainable}, DRA~\cite{ding2022catching}, and AHL~\cite{zhu2024anomaly}. MLEP, DevNet, DRA, and AHL are specifically designed for OSAD. SAOE is a supervised detector enhanced with synthetic anomalies and anomaly exposure, whereas FLOS is an imbalanced classifier leveraging focal loss.
		
		\subsection{Implementation Details}
		The input image size is $448 \times 448 \times 3$. We set  $K$ in the top-$K$ MIL to 10\% of the number of all scores per score map. AdamW optimizer~\cite{loshchilov2017decoupled} is used for the parameter optimization using an initial learning rate $2\times 10^{-4}$ with a weight decay of $1 \times 10^{-5}$. DPDL is trained on one NVIDIA GeForce RTX $4090$ GPU, which are trained using 50 epochs, with 20 iterations per epoch. Following previous protocol~\cite{ding2022catching,zhu2024anomaly}, we evaluate performance with anomaly sample numbers of \( M = 10 \) and \( M = 1 \), and for robust detection of unseen anomalies, we use CutMix~\cite{yun2019cutmix} to create pseudo-anomaly samples as augmented data for known anomalies. The prototype quantity $C$ is set to 32 as default.  
		\begin{table}[!ht]
			\captionsetup{skip=3pt}
			\caption{An ablation study for $\mathsf{M}_\text{n}$, $\mathsf{M}_\text{a}$ and $\mathsf{M}_\text{r}$.}
			\centering
			\scalebox{0.88}{
				\setlength{\tabcolsep}{4pt} 
				\begin{tabular}{ccc|cccc}
					\toprule
					\multirow{1}{*}{$\mathsf{M}_\text{n}$} &
					\multirow{1}{*}{$\mathsf{M}_\text{a}$} &
					\multirow{1}{*}{$\mathsf{M}_\text{r}$} &
					\multicolumn{1}{c}{AITEX} & \multicolumn{1}{c}{ELPV} & \multicolumn{1}{c}{Mastcam} \\
					\midrule
					\multicolumn{6}{c}{Ten Training Anomaly Examples Under General Settings} \\
					\midrule
					$\usym{2717}$ & $\usym{2717}$ & $\usym{2713}$ & 0.928 $\pm$ 0.019 & 0.914 $\pm$ 0.021 & 0.899 $\pm$ 0.036 \\
					$\usym{2713}$ & $\usym{2717}$ & $\usym{2713}$ & 0.939 $\pm$ 0.023 & 0.919 $\pm$ 0.011 & 0.908 $\pm$ 0.017 \\
					$\usym{2717}$ & $\usym{2713}$ & $\usym{2713}$ &\underline{0.963} $\pm$ 0.008 & \underline{0.930} $\pm$ 0.013 & \underline{0.924} $\pm$ 0.019 \\
					$\usym{2713}$ & $\usym{2713}$ & $\usym{2713}$ &\textbf{0.975} $\pm$ 0.007 & \textbf{0.937} $\pm$ 0.003 & \textbf{0.934} $\pm$ 0.010 \\
					\midrule
					\multicolumn{6}{c}{Ten Training Anomaly Examples Under Hard Settings} \\
					\midrule
					$\usym{2717}$ & $\usym{2717}$ & $\usym{2713}$ & 0.746 $\pm$ 0.025  & 0.797  $\pm$ 0.026 & 0.723 $\pm$ 0.017  \\
					$\usym{2713}$ & $\usym{2717}$ & $\usym{2713}$  & 0.758 $\pm$ 0.035 &  0.802 $\pm$ 0.024 & 0.736 $\pm$ 0.028 \\
					$\usym{2717}$ & $\usym{2713}$ & $\usym{2713}$ & \underline{0.781} $\pm$ 0.014 & \underline{0.811} $\pm$ 0.018 & \underline{0.762} $\pm$ 0.021 \\
					$\usym{2713}$ & $\usym{2713}$ & $\usym{2713}$ & \textbf{0.798} $\pm$ 0.005 & \textbf{0.818} $\pm$ 0.003 & \textbf{0.778} $\pm$ 0.007 \\
					\bottomrule
			\end{tabular}}
			\label{tab:ablation}
			\vspace{-15pt}
		\end{table}
		
		\subsection{Results under General Setting}
		Tab.~\ref{tab:main_performance} highlights DPDL’s strong performance. In the challenging scenario of single-anomaly detection, DPDL improves the performance of AHL~\cite{zhu2024anomaly} by more than 8.3\% on the AITEX, ELPV, and Mastcam datasets. Furthermore, it achieves significant improvements across six additional datasets, which suggest the effective utilization of few-shot anomaly examples in DPDL, while mitigating overfitting to the seen anomalies.  When shifting to ten anomaly examples settings, DPDL continues to maintain a significant lead with over 5.4\% improvement on those datasets. Given the rich and diverse set of normal samples in these datasets, DPDL leverages the DPL component to encapsulate these samples within a compact, discriminative distribution space, while effectively pushing anomalous samples outside this space, thereby enabling accurate anomaly detection. In the setting with ten abnormal samples, although existing methods have reached performance saturation on the MVTecAD, Optical, and SDD datasets, DPDL still has a certain lead, demonstrating the strong ability of DPL and DFL in learning tight boundaries of normal sample distributions and generalizing to previously unseen anomaly domains. Furthermore, when evaluated on the medical datasets BrainMRI and HeadCT, DPDL demonstrates competitive performance despite these datasets being notably small in scale and containing only a single class. This highlights the algorithm's ability to deliver robust results even in data-scarce conditions.

		\subsection{Results under the Hard Setting}
		Tab.~\ref{tab:hard_setting} summarizes the performance comparison under the hard setting.  It is evident that DPDL achieves the highest AUC scores in both the single-anomaly and ten-anomaly sample settings. Specifically, compared to the closest competing method, AHL~\cite{zhu2024anomaly}, DPDL achieves an improvement in AUC scores ranging from 0.6\% to 7.9\%  in the ten anomaly examples settings and from 0.5\% to 8.9\% in the one anomaly example settings, respectively.
		The observed improvement can be attributed to the strong generalization ability of DPDL in detecting unseen anomaly classes, even when the model is trained on only a single anomaly class.
		\begin{figure*}[htbp]
			\centering
			\includegraphics[width=0.99\textwidth]{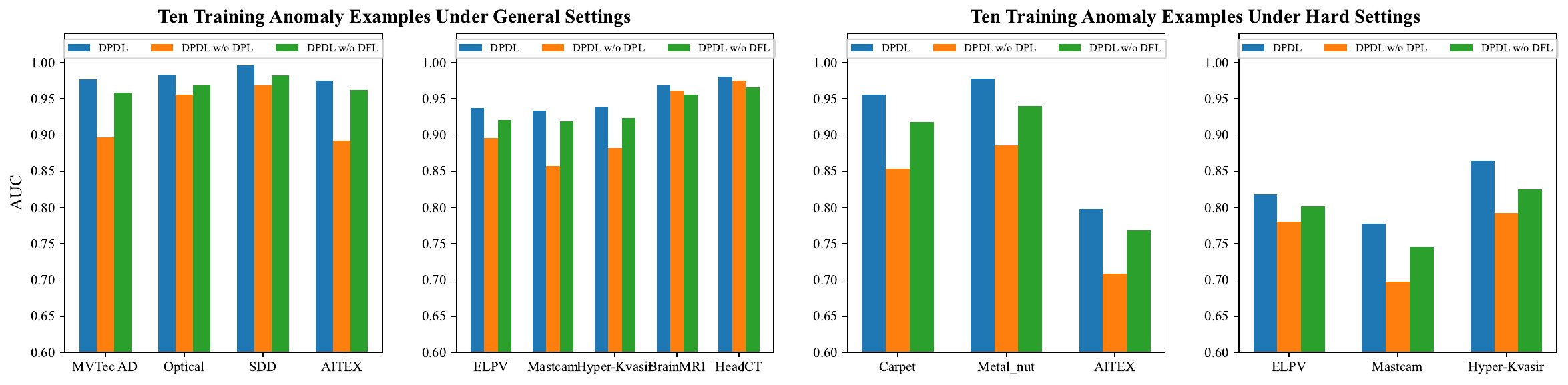}
			\captionsetup{skip=0.2pt}
			\caption{Ablation study for SB and DFL  under the general settings and hard settings.}
			\label{fig:ablation}
		\end{figure*}
		\begin{figure*}[htbp]
			\centering
			\includegraphics[width=0.99\textwidth]{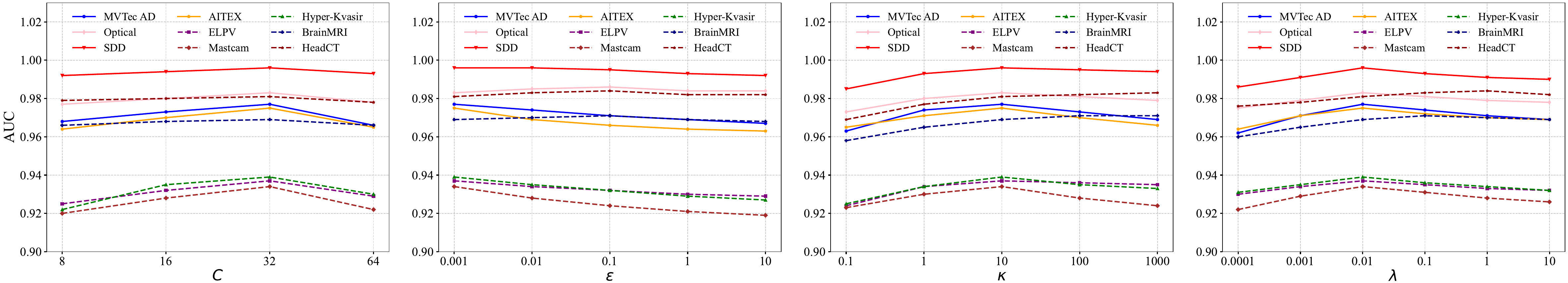}
			\captionsetup{skip=0.2pt}
			\caption{Parameter sensitivity analysis  for $C$, $\epsilon$, $\kappa$ and $\lambda$.}
			\label{fig:sensitivity}
						\vspace{-15pt}
		\end{figure*}

		\subsection{Ablation Study}
		The ablation study in Fig.~\ref{fig:ablation}  highlights the critical roles of the DPL and DFL components in improving DPDL's open-set anomaly detection. We denote the variants that remove only DPL or DFL as `DPDL w/o DPL' and `DPDL w/o DFL', respectively. Compared to the full DPDL model, removing neither DPL nor DFL leads to a significant AUC decline, illustrating their critical utility. Specifically, `DPDL w/o DPL' exhibits the most significant performance drop on industrial anomaly datasets, reflecting DPL's prominent role in learning precise and tight distribution boundaries for normal samples. By transforming normal samples into Gaussian distribution prototype space and pushing abnormal samples away, DPL enhances the recognition ability of anomaly samples. Meanwhile, ablation studies on `DPDL w/o DFL' further highlight the critical role of the DFL component. By performing discreteness feature learning in hyperspherical space, DFL enhances the generalization ability to out-of-distribution anomalies.
		
		Additionally, the ablation experiments on \(\mathsf{M}_\text{n}\), \(\mathsf{M}_\text{a}\), and \(\mathsf{M}_\text{r}\) across three datasets in Tab.~\ref{tab:ablation} reveal their varying contributions.  DPDL shows the most significant performance drop when \(\mathsf{M}_\text{r}\) is removed, which illustrates that \(\mathsf{M}_\text{r}\)  plays the most critical role in detecting anomalies. Furthermore, the removal of $\mathsf{M}_\text{a}$ and $\mathsf{M}_\text{n}$ lead to a noticeable performance drop in DPDL, which underscores their essential roles.
		
		\subsection{Parameter Sensitivity Analysis}
		Fig.~\ref{fig:sensitivity} illustrates the results of the four hyperparameters under the general settings across nine datasets. Overall, the performance remains stable within a certain range of hyper-parameter variations, demonstrating the 
		DPDL's robustness. 
		
		\noindent \textbf{Prototype quantity $C$.} We begin by investigating the critical impact of the number of initialized prototypes $C$ on distributed prototype learning in DPDL. We select $\{8, 16, 32, 64\}$ as the values for the hyperparameters. As $C$ increases, DPDL's performance improves steadily, but excessively large values of $C$ hinder the model's effectiveness. This phenomenon is particularly pronounced on AITEX, Mastcam, MVTecAD, and Hyper-Kvasir, which contain more categories. One possible explanation is that a prototype space with too few prototypes loses discriminative information, while an excessively large number of prototypes reduces the compactness of the space.
		
		\noindent \textbf{DPL trajectory $\epsilon$ in Eqns.~(\ref{eqn:eq9}), (\ref{eqn:eq10}), (\ref{eqn:eq3}) and (\ref{eqn:g-gmm}).}  It can be observed that, particularly on the AITEX, Mastcam, and MVTecAD datasets with a larger number of categories, DPDL exhibits a relatively stable performance decline as  \( \epsilon\) increases. As a crucial parameter in SB,  $\epsilon$ governs the trajectory state. Since smaller values produce straighter trajectories and larger values increase fluctuation, smaller \( \epsilon\) facilitates sampling more robust abstract prototypes from the relatively dispersed conditional distribution. 
		
		\noindent \textbf{DFL tightness $\kappa$.} According to  Fig.~\ref{fig:sensitivity}, increasing \( \kappa \) generally enhances model performance, but values above \( \kappa = 10 \) introduce negative effects in certain scenarios. A possible reason is that excessive sample dispersion makes it more challenging to tighten the normal distribution boundary. 
		
		\noindent \textbf{Loss parameter $\lambda$.}
		We conduct a sensitivity analysis on the loss parameters \( \lambda \). It can be observe that setting \( \lambda = 0.01 \) achieves optimal performance on seven larger-scale datasets, while \( \lambda = 1 \) yields the best results on two datasets with limited data. As \( \lambda \) increases, the performance declines, potentially due to gradient conflicts among the dispersion loss, SB transform loss and the main task loss.

		\section{Conclusion}

		We propose Distribution Prototype Diffusion Learning (DPDL) for OSAD. DPDL leverages schrödinger bridge to map the normal distribution to a prototype space, simultaneously repelling anomalies to facilitate precise anomaly detection. We propose a dispersion feature learning way in hyperspherical space, which benefits the detection of out-of-distribution anomalies. Experimental results illustrate DPDL's robustness in diverse anomaly detection scenarios.

		
		{
			\small
			\bibliographystyle{ieeenat_fullname}
			\bibliography{main}
		}
		
			
			\end{document}